\documentclass{article} 
\usepackage{fancyhdr}

\usepackage[accepted]{aistats2016}

\usepackage{mathptmx}
\usepackage{amsmath}
\usepackage{amsthm}
\usepackage{amssymb}

\usepackage{color,soul}

\usepackage{graphicx}                               
\usepackage{amsmath}                                
\usepackage{natbib}                   

\definecolor{royalblue}{rgb}{0.0, 0.14, 0.4}
\usepackage[colorlinks]{hyperref}
\hypersetup{citecolor=royalblue}
\hypersetup{linkcolor=royalblue}

\DeclareRobustCommand{\parhead}[1]{\textbf{#1} }

\newcommand{\E}{\mathbb{E}}

\newcommand{\cL}{\mathcal{L}}

\usepackage{url}

\newcommand\g[1][]{\:#1\vert\:}

\newcommand{\bx}{\mathbf{x}}

\newcommand{\bz}{\mathbf{z}}

\usepackage{algorithm}
\usepackage[noend]{algpseudocode}

\newcommand{\be}{\begin{eqnarray}}
\newcommand{\ee}{\end{eqnarray}}
\newcommand{\n}{\nonumber \\}
\newcommand{\tinv}{ T^{-1} }

\begin{document}

\twocolumn[
\aistatstitle{Variational Tempering}
\aistatsauthor{
Stephan Mandt \And James McInerney 
}
\aistatsaddress{
Columbia University \\ \texttt{sm3976@columbia.edu} \And Columbia University \\ \texttt{jm4181@columbia.edu}
}
\aistatsauthor{
Farhan Abrol  \And Rajesh Ranganath \And David Blei
}
\aistatsaddress{
Princeton University \\ \texttt{fabrol@cs.princeton.edu} \And Princeton University \\ \texttt{rajeshr@cs.princeton.edu} \And Columbia University \\ \texttt{david.blei@columbia.edu} 
}
\runningauthor{Stephan Mandt, James McInerney, Farhan Abrol, Rajesh Ranganath, David Blei}  
\runningtitle{Variational Tempering} 
]


\begin{abstract}

  Variational inference (VI) combined with data subsampling
  enables approximate posterior
  inference over large data sets, but
  suffers from poor local
  optima. We first formulate a deterministic annealing approach for
  the generic class of conditionally conjugate exponential family models.
  This approach uses a decreasing temperature parameter which deterministically
  deforms the objective during the course of the optimization. A well-known drawback to this
  annealing approach is the choice of the cooling schedule. 
  We therefore introduce variational tempering, a variational
  algorithm that introduces a temperature latent variable to the model. 
  In contrast to related work in the Markov chain Monte Carlo literature, this algorithm 
  results in adaptive annealing schedules. Lastly, we
  develop local variational tempering, which
  assigns a latent temperature to each data point; this
  allows for dynamic annealing that varies across data.
  Compared to the traditional VI, all proposed
  approaches find improved predictive likelihoods on held-out data.

\end{abstract}

\section{Introduction}

Annealing is an ancient metallurgical practice.  To form a tool,
blacksmiths would heat a metal, maintain it at a suitable temperature,
and then slowly cool it.  This relieved its internal stresses, made it
more malleable, and ultimately more workable.  
We can interpret this process as an
optimization with better outcomes when temperature is annealed.

The physical process of annealing has analogies in non-convex
optimization, where the cooling process is mimicked in different ways.
Deterministic annealing~\citep{rose1990deterministic}
uses a temperature parameter to deterministically deform the objective
according to a time-dependent schedule.  The goal is for the
parameterized deformation to smooth out the objective function and
prevent the optimization from getting stuck in shallow local optima.  

Variational inference turns posterior inference
into a non-convex optimization problem, one whose objective has many
local optima. We will explore different approaches based on annealing as a way 
to avoid some of these local optima. Intuitively, the variational objective trades off 
variational distributions that fit
the data with variational distributions that have high
entropy. Annealing penalizes the low-entropy distributions and then
slowly relaxes this penalty to give more weight to distributions that
better fit the data.  

We first formulate deterministic annealing for stochastic variational
inference (SVI), a scalable algorithm for finding approximate
posteriors in a large class of probabilistic
models~\citep{bleistochastic}. 
Annealing necessitates the manual construction and search over 
temperature schedules, a computationally
expensive procedure. To sidestep having to set the
temperature schedule, we propose two methods that treat the
temperature as an auxiliary random variable in the model.  Performing
inference on this expanded model---which we call variational tempering (VT)---allows us to use the data to automatically
infer a good temperature schedule. 
We finally introduce local variational tempering (LVT), an algorithm
that assigns different temperatures to individual data points and thereby simultaneously anneals at many different rates.

We apply deterministic annealing and variational tempering 
to latent Dirichlet allocation, a topic model~\citep{blei2003latent}, and test it on
three large text corpora involving millions of documents. 
Additionally, we study the factorial mixture model~\citep{ghahramani1995factorial} with both artificial
data and image data. 
We find that deterministic annealing 
finds higher likelihoods on held-out data than  
stochastic variational inference. We also find that in all  cases, variational tempering performs as well or better than
 the optimal annealing schedule, eliminating the need for temperature parameter search and opening paths to automated annealing.

\parhead{Related work to annealing.}  The roots of annealing reach back
to \citet{metropolis1953equation} and to
\citet{kirkpatrick1983optimization}, where the objective is 
corrupted through
the introduction of temperature-dependent noise.
Deterministic annealing was originally
used for data clustering applications
\citep{rose1990deterministic}. It was later applied to latent variable
models in \citet{uedanakano}, who suggest deterministic annealing for
maximum likelihood estimation with incomplete data, and specific Bayesian
models such as latent factor models~\citep{ghahramani2000variational}, hidden Markov models~\citep{Katahira:2008} 
and sparse factor models~\citep{yoshidabayesian}. Generalizing these model-specific approaches,
we formulate annealing for the
general class of conditionally conjugate exponential family models and compare it
to our tempering approach that learns the temperatures from the data. 
In contrast to earlier works, we also combine deterministic annealing with stochastic variational inference, scaling it up to massive data sets.

\parhead{Related work to variational tempering.}
VT is inspired by
multicanonical Monte Carlo methods. 
These Markov chain Monte Carlo algorithms
sample at many temperatures,
thereby enhancing mixing times of the Markov chain~\citep{swendsen1986replica,
 geyer1991markov, berg1992multicanonical, marinari1992simulated}. 
Our VT approach introduces global auxiliary temperatures in a similar way, 
but for a variational algorithm there is no notion of a mixing time. 
Instead, the key idea is that the variational algorithm learns a distribution over temperatures from the data,
and that the expected temperatures adjust the statistical weight of each update over iterations.
Our LVT algorithm is different in that the temperature variables
are defined per data point.

\section{Annealed Variational Inference and Variational Tempering}

Variational tempering (VT) and local variational tempering (LVT) 
are extensions of annealed variational inference (AVI). All three algorithms
are based on artificial temperatures and are introduced in a common theoretical framework.
We first give background about mean-field variational inference.  We
then describe the modified objective functions and algorithms for
optimizing them.  We embed these methods into stochastic
variational inference~\citep{bleistochastic}, optimizing the variational 
objectives over massive data sets.

\subsection{Background: Mean-Field Variational Inference}
\label{sec:method_vi}

We consider hierarchical Bayesian models. In these models the global variables are shared across data points
and each data point has a local hidden variable.  Let $\bx = x_{1:N}$
be observations, $\bz = z_{1:N}$ be local hidden variables, and
$\beta$ be global hidden variables. We define the model by the joint,
\begin{align*}
  p(\beta, \bz, \bx) = p(\beta \g \alpha) \textstyle \prod_{i=1}^{N} p(z_i, x_i \g \beta), 
\end{align*}
where $\alpha$ are hyperparameters for the global hidden variables.
Many  machine learning models have this form~\citep{bleistochastic}.

The main computational problem for Bayesian modeling is posterior
inference.  The goal is to compute
$p(\beta, \bz \g \bx)$, the conditional distribution of the latent
variables given the observations.  For many models this
calculation is intractable, and we must resort to
approximate solutions.

Variational inference proposes a parameterized family of distributions
over the hidden variables $q(\beta, \bz \g \nu)$ and tries to find the
member of the family that is closest in KL divergence to the posterior~\citep{JordanVariational}.
This is equivalent to optimizing the
evidence lower bound (ELBO) with respect to the variational
parameters,
\be
  \mathcal{L}(\nu)  = \E_q[{\log p(\beta, \bz, \bx)}] - \E_q[\log q(\beta, \bz \g \nu)] .\label{eq:elbo}  
\ee
Mean-field variational inference uses the fully factorized family,
where each hidden variable is independent,
\begin{align*}
  q(\beta, z \g \nu) = q(\beta \g \lambda) \textstyle \prod_{i=1}^{N} q(z_i \g \phi_i).
\end{align*}
The variational parameters are $\nu = \{\lambda, \phi_{1:N}\}$, where
$\lambda$ are global variational parameters and $\phi_i$
are local variational parameters. Variational inference algorithms optimize 
Eq.~\ref{eq:elbo} with coordinate or gradient ascent.
This objective is non-convex.  
To find better local optima, we use AVI and VT.

\subsection{Annealed Variational Inference (AVI)}
\label{sec:method_tempered_posterior}

AVI applies deterministic annealing to
mean-field variational inference.  To begin, we introduce a
temperature parameter $T\geq 1$. Given $T$, we define a
joint distribution as
\be
  \label{eq:tempered-joint}
  p(\beta, \bz, \bx \g T) = \frac{p(\bz, \bx \g \beta)^{1/T} p(\beta \g \alpha)}{C(T)},
\ee
where $C(T)$ is the normalizing constant.  
We call $p(\beta, \bz, \bx \g T)$ the \textit{annealed joint}. 
In contrast to earlier work, we anneal the likelihoods instead of the posterior~\citep{ghahramani2000variational,
Katahira:2008, yoshidabayesian}, which we will comment on later in this subsection.
Note that setting $T=1$ recovers the original model.

The normalizing constant, called the \emph{tempered partition function}, integrates out the joint,
\be
  \label{eq:tempered-normalizer}
  C(T) = \int p(\bz, \bx \g \beta)^{1/T} p(\beta) d\bx d\bz d\beta.
\ee
For AVI, we do not need to calculate the tempered partition function as constant terms do not affect the variational objective. 
For VT, we need to approximate this quantity (see Section~\ref{sec:updates_VT}). 

The annealed joint implies the annealed posterior.
AVI optimizes the
variational distribution $q(\cdot)$ against a sequence of annealed
posteriors.  We begin with high temperatures and end in the original
posterior, i.e., $T=1$. 
In more detail, at each stage of annealed variational inference we fix
the temperature $T$.  We then (partially) optimize the mean-field ELBO
of Eq.~\ref{eq:elbo} applied to the annealed model of
Eq.~\ref{eq:tempered-joint}.  We call this the annealed ELBO,
\be
  \label{eq:temperedelbo}
  \begin{split}
    \cL_A(\lambda,\phi; T) \; =\;  \E_{q}[\log p(\beta \g \alpha)] -  \E_{q}[\log q(\beta \g \lambda)] \\
     \quad + \textstyle \sum_{i=1}^N \left( \E_{q}[\log p(x_i,z_i \g
      \beta)]/T - \E_{q}[\log q(z_i \g \phi_i)] \right). 
  \end{split}
\ee
We then lower the temperature. We repeat until we reach $T=1$ and the optimization has converged.

As expected, when $T=1$ the annealed ELBO is the traditional ELBO.
Note that the annealed ELBO does not require the normalizer $C(T)$
in Eq.~\ref{eq:tempered-normalizer} because it does not depend on any of
the latent variables.

Why does annealing work?  The first and third terms on the right hand side of
Eq.~\ref{eq:temperedelbo} are the expected log prior and the log
likelihood, respectively.  Maximizing those terms with respect to $q$ causes the approximation to place its
probability mass on configurations of the hidden variables that best
explain the observations; this induces a rugged objective with many
local optima.  The second and fourth terms together are the entropy of the
variational distribution.  The entropy is concave: it acts like a regularizer
that prefers the variational distribution to be spread across
configurations of the hidden variables. By first downweighting the likelihood by $1/T$, 
we favor smooth and entropic distributions. By gradually lowering $T$
we ask the variational distribution to put more weight on explaining the data.

\begin{figure} 
\begin{center}
  \includegraphics[width=0.49\linewidth]{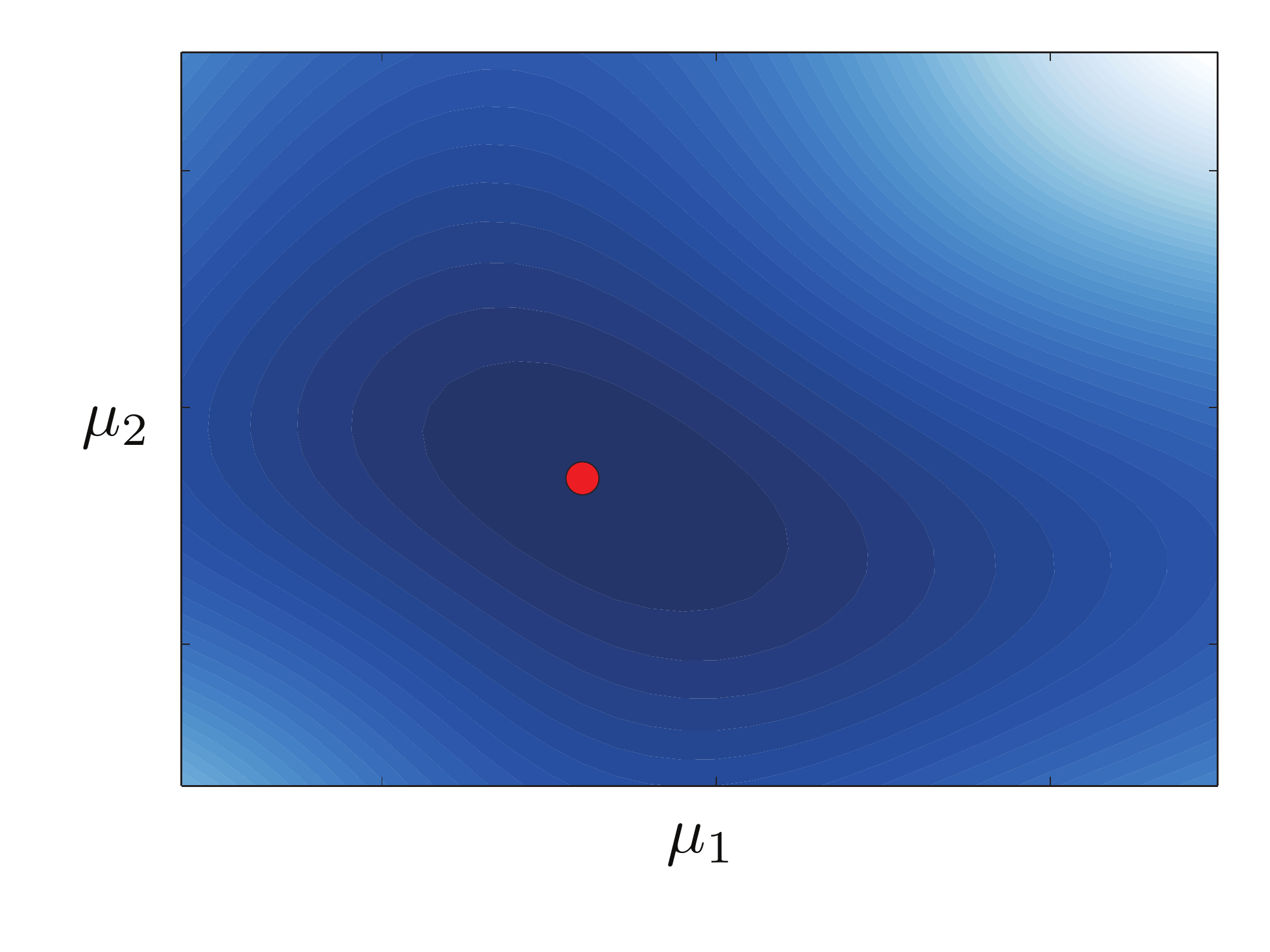}
  \includegraphics[width=0.49\linewidth]{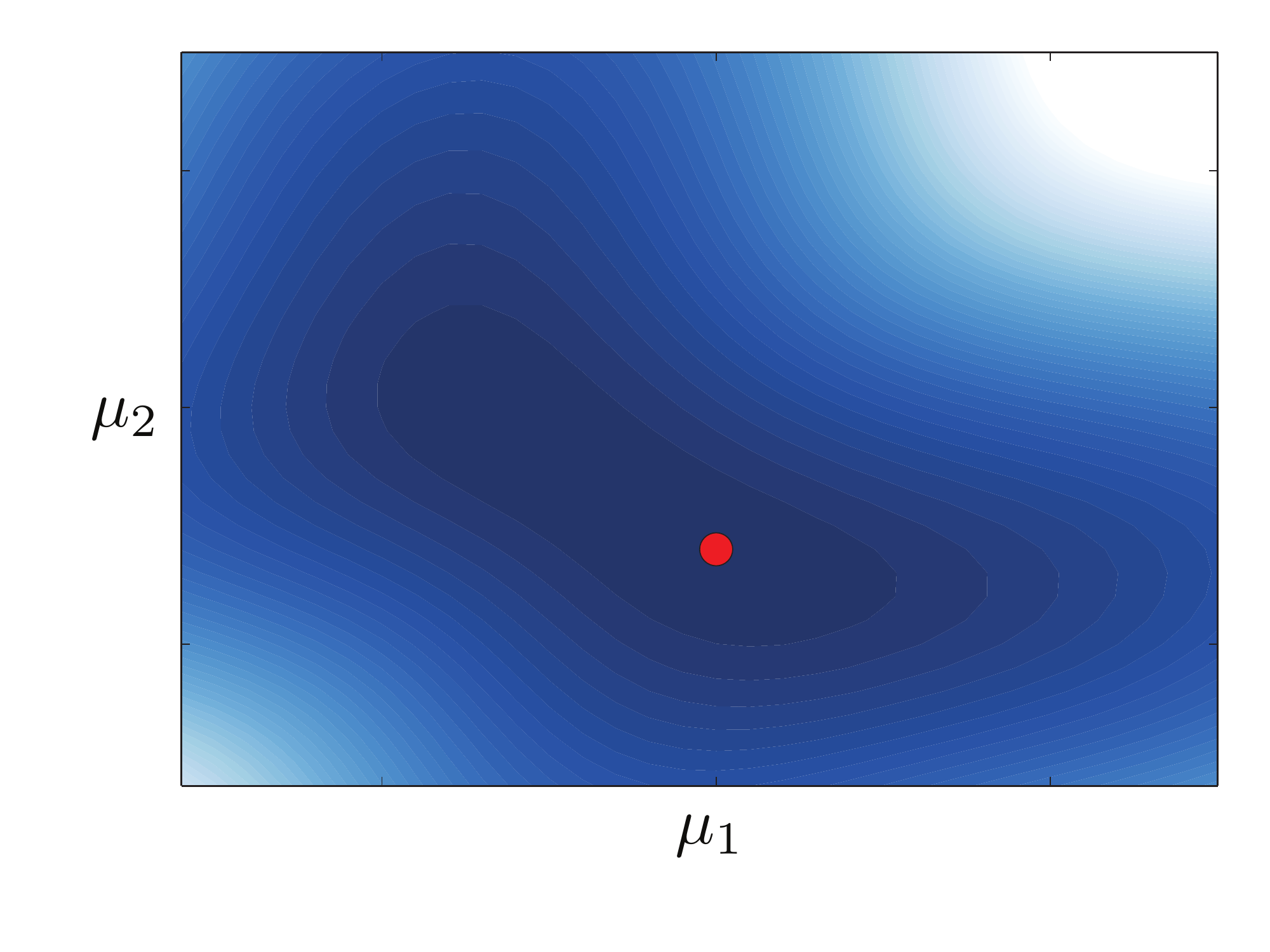}
  \includegraphics[width=0.49\linewidth]{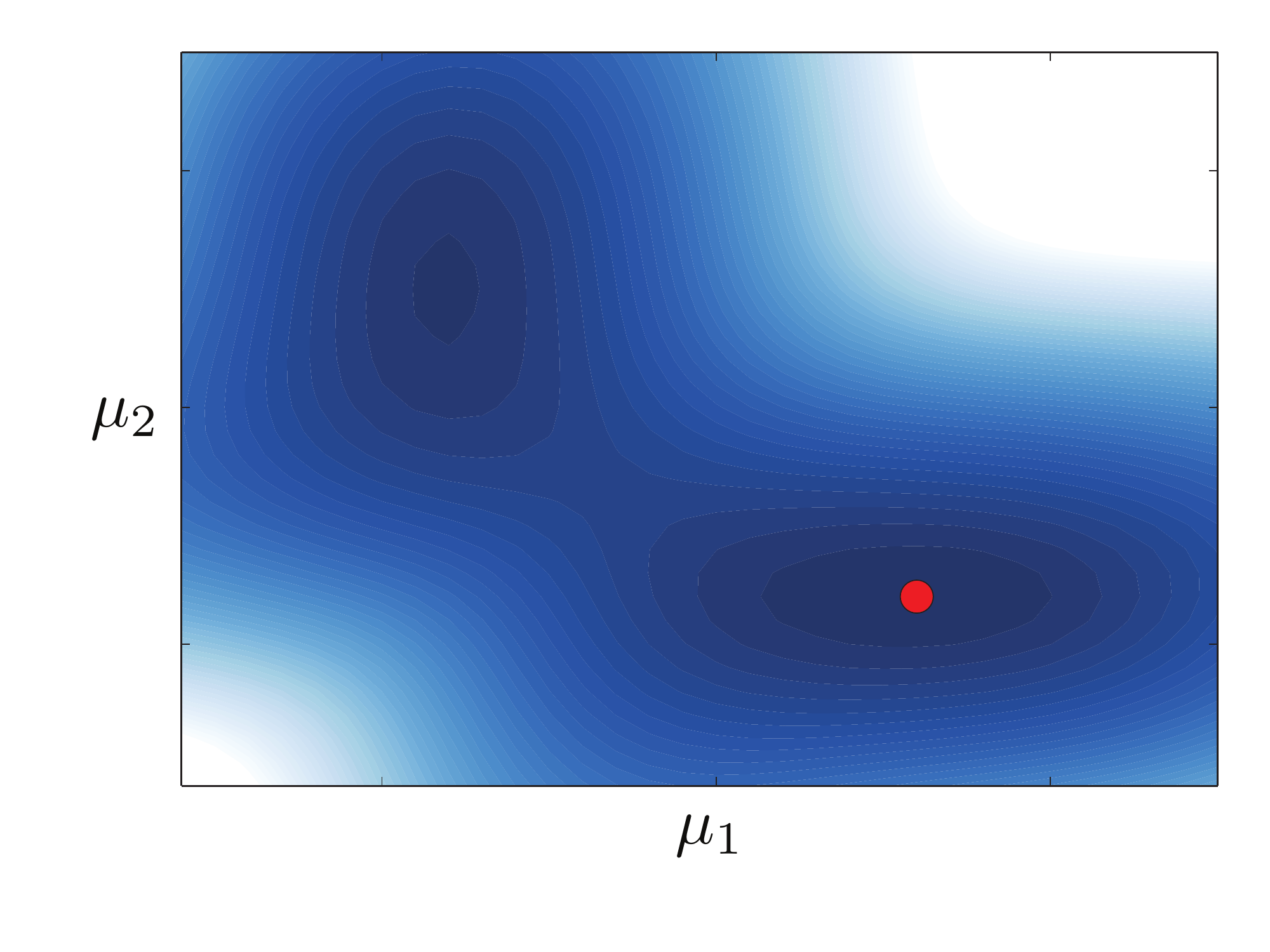}
  \includegraphics[width=0.49\linewidth]{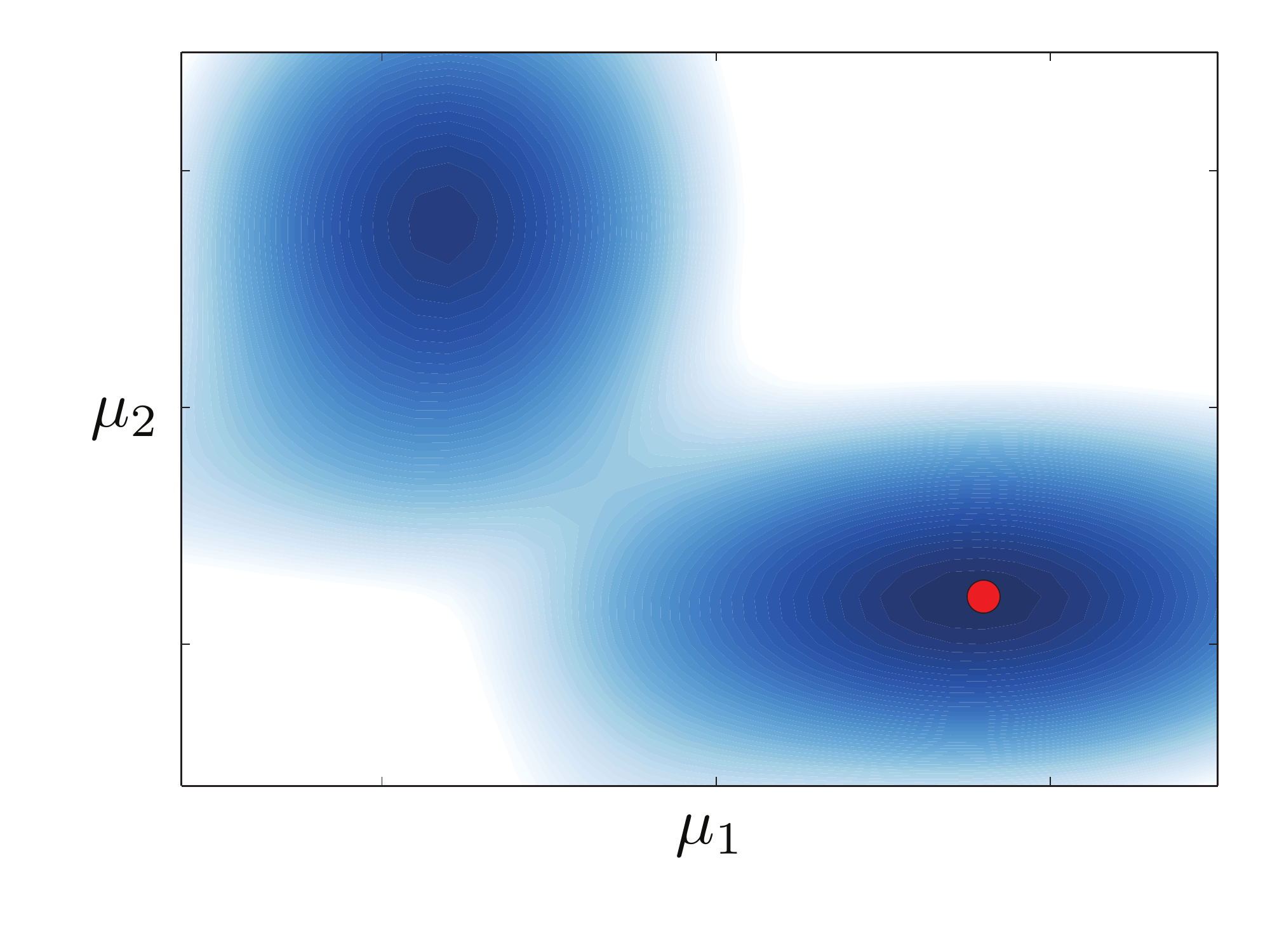}
\end{center}
  \caption{AVI for a mixture model of two
    Gaussians.  We show the variational objective as a function of the
    two latent Gaussian means $\mu_0,\mu_1$ for temperatures (left to
    right) $T = 20$, $T=13$, $T=8.5$ and $T=1$. The red dot indicates the global optimum.}
\label{fig:contour}
\end{figure}

Fig.~\ref{fig:contour} shows the annealed ELBO for a mixture of
two one-dimensional Gaussians, also discussed in~\cite{Katahira:2008}.  
At large $T$, the objective has a single
optimum and the algorithm moves to a good region of the objective.
The decreasing temperature reveals the two local optima. Thanks to
annealing, the algorithm is positioned to move to the better (i.e., more global) optimum. 

We noted before that in other formulations of annealing one typically defines the annealed
posterior to be $p(\bz, \beta \g \bx,T) \propto p(\bz, \beta | \bx)^{1/T}$,
which anneals both the likelihood and the
prior~\citep{neal1993probabilistic}.  This approach has nearly the same
affect but can lead to practical problems. The temperature affects the prior on the global variables, which can lead to extremely skewed priors
 that cause the gradient to get stuck in early iterations.
As an example, consider the gamma distribution with shape$=0.05$.  Annealing
this distribution with $T=2$ reduces the 50th percentile of this distribution
by over five orders of magnitude. By only annealing the likelihood and leaving the
prior fixed this problem does not occur.

\parhead{Interplay of annealing and learning schedules.}
Our paper treats annealing in a gradient-based setup.
Coordinate ascent-based annealing is simpler because there is no learning schedule and therefore
the slower we anneal, the better we can track good optima 
(as can be seen in Fig.~\ref{fig:contour}). 
In contrast, when following gradients, 
the temperature schedule and the learning rate schedule become intertwined. 
When we anneal too slowly, the gradient descent algorithm may approximately
converge before the annealed objective has reached its final shape. This leads to 
suboptimal solutions (e.g., we found in experiments with latent Dirichlet allocation that annealing performs worse than SVI 
in many cases when temperature is larger than 5). This makes finding good temperature schedules in gradient-based variational annealing hard in practice (finding appropriate learning rates also is hard by itself~\citep{adaptive,duchi2011adaptive}).

\subsection{Variational Tempering (VT)}
\label{sec:method_t_rv}
\label{sec:updates_tempered_model}

Since finding an
appropriate schedule can be difficult in practice, we focus on 
adaptive annealing schedules that learn
a sequence of temperatures from the data.
We build on AVI to develop VT, a method that
learns a temperature schedule from the data. 
VT introduces temperature as an auxiliary variable in the model; it recovers the
original model (and thus original posterior) when the
temperature is equal to 1.

\parhead{Random temperatures.} 
As was the case with AVI, 
VT relies on physical heuristics. 
It mimics a physical system where several temperatures coexist at the same time,
hence where temperature is a random variable that has a joint distribution
with the other variables in the system. 
We consider a finite set of possible
temperatures,
\begin{align*}
 1\equiv T_1 < T_2 < \cdots < T_M.
\end{align*}
A finite, discrete set of
temperatures is convenient as it allows us to precompute the tempered
partition functions, each temperature leading to a Monte Carlo integral
whose dimension does not depend on the data (this is described in Section~\ref{sec:updates_VT}).

\parhead{Multinomial temperature assignments.}
We introduce a  random variable that
assigns joint distributions to temperatures. Conditional on the
outcome of that random variable, the model is
an annealed joint distribution at that temperature.
We define a multinomial temperature assignment,
\begin{align*}
y \sim \textrm{Mult}(\pi).
\end{align*}
We treat $\pi$ as fixed parameters
and typically set $\pi_m = 1/M$.

\parhead{Tempered joint.} 
The joint distribution factorizes as $p(\bx,\bz,\beta,y)  =  p(\bx,\bz,\beta|y) p(y)$.
We place a uniform prior over temperature assignments,
$p(y) =  \prod_{m=1}^M\pi_m^{y_{m}}$. Conditioned on $y$, we define the
tempered joint distribution as
\begin{align*}
p(\bx,\bz,\beta|y)  =  p(\beta) \frac{1}{C(T_y)} \prod_{i=1}^N p(x_i,z_i | \beta)^{1/T_y} .
\end{align*}
This allows us to define the model as
\begin{align*}
p(\bx,\bz,\beta,y) =  p(\beta)\prod_{m=1}^M \left(\frac{ \pi_m}{C(T_m)}  \prod_{i=1}^N p(x_i,z_i |\beta)^{1/T_{m}} \right)^{y_{m}}.
\end{align*}

\parhead{The tempered ELBO.}

We now define the variational objective for the expanded
 model. We extend the mean-field
family to contain a factor for the
temperature,
\begin{align*}
  q(\bz,\beta,y \g \phi, \lambda, r) = q(\bz|\phi)
  q(\beta|\lambda) q(y|r),
\end{align*}
where we introduced a variational multinomial for the temperature with variational parameter $r$,
$q(y|r) = \prod_{m=1}^M r_m^{y_{m}}$.

Using this family, we augment the annealed ELBO.  It now contains
terms for the random temperature and explicitly includes $\log C(T)$.
The tempered evidence lower bound (T-ELBO) $ \mathcal{L}_T \equiv  \mathcal{L}_T (\lambda,\phi,r)$ is
\be
 \mathcal{L}_T   &=&   \E_{q}[\log p(\beta)]+ \E_{q}[\log p(y)] - {\mathbb E}_{q}[\log q(\beta) ]  \n
 & & +  \E_q[1/T_y]  {\textstyle \sum_{i}}   \E_{q}[ \log p(x_i,z_i|\beta)] \label{eq:telbo} \\
 & &- \E_q[  \log C(T_{y})]  -  {\textstyle \sum_{i}} {\mathbb E}_{q}[\log q(z_i) ] -  \E_{q}[\log q(y)]. \nonumber
\ee
When comparing the T-ELBO with the annealed ELBO in Eq.~\ref{eq:temperedelbo}, we see that
\emph{expected} local inverse temperatures $\E_q[1/T_{y}]  \equiv {\textstyle \sum_{m}^{M}} \E_q[y_{m}/T_m] $ 
in VT play the role of the (global) inverse temperature parameter in
AVI. As these expected temperatures typically decrease during learning (as we show),
the remaining parts of the tempered ELBO  will effectively be annealed over iterations.
In VT, we optimize this lower bound to obtain a variational approximation of the
posterior.

\parhead{The tempered partition function.}
We will now comment on the role of $\log C(T_y)$ which appears in the T-ELBO (see Eq.~\ref{eq:telbo}). In contrast to annealing, we cannot omit this term.

Without $C(T_y)$, the model would place all its probability mass
for $m$ around the highest possible temperature $T_{M}$.  To see this, note
that log likelihoods are generally negative, thus
$\E_q[\log p(\bx,\bz|\beta)]<0$. If we did not add $- \log C(T_y)$ to the T-ELBO, 
maximizing the objective would require us to minimize $\E[1/T_{y}]$. 

The log partition function in the T-ELBO prevents temperatures from
taking their maximum value. It is
usually a monotonically increasing function in $T$.  This way 
$\log C(T_y)$ penalizes large values of T. 

\subsection{Local Variational Tempering (LVT)}
Instead of working with global temperatures, which are shared across data points,
we can define temperatures unique to each data point. This allows us to better fit the
data and to learn annealing schedules specific to individual data points.
We call this approach local variational tempering (LVT).

In this approach, $T_i$ is a per-data point local temperature, and we define the tempered joint as follows:
\be
p(\bx ,\bz ,\beta,{\bf T})  \propto    p(\beta) \prod_{i=1}^N \left[p(x_i,z_i|\beta)^{1/T_i}p(T_i) \right].\label{eq:LVI}
\ee
In the global case, the temperature distribution was limited to discrete support due to normalization. Here we have more flexibility,
$p(T_i)$ can have discrete (multinomial) or continuous support (e.g. $1/T$ could be beta-distributed).

The model can also be formulated as
$
p(\bx ,\bz ,\beta,{\bf T}) =  p(\beta) \prod_{i=1}^N p(x_i,z_i|\beta/T_i)p(T_i),
$
where we see that temperature downweights the global hidden variables and therefore makes the local conditional distributions
more entropic. The advantage of this formulation is that we do not have to compute the tempered partition function as
the local-tempered likelihood is in the same family as the original one (Section~\ref{sec:SAVI}). 
On the downside, the resulting model is non-conjugate.

Local temperature describes the likelihood that a particular data point came from the non-tempered model.
Outliers can be better explained in this model by assigning them a high local temperature. 
$T_i$ therefore allows us to more flexibly model the data. 
It also enables us to learn a different annealing schedule for each data point during inference. 

\section{Algorithms for Variational Tempering}
\label{sec:SAVI}

We now introduce the annealing and variational tempering algorithms
for the general class of local and global hidden variables discussed in Section~\ref{sec:method_vi}.
Our  algorithms are based on stochastic variational inference~\citep{bleistochastic}, a scalable 
Bayesian inference algorithm that uses stochastic optimization.

\subsection{Conditionally Conjugate Exponential Families}
As in~\citep{bleistochastic}, we focus on the conditionally conjugate
exponential family (CCEF).  A model is a CCEF if the prior and local
conditional are both in the exponential family and form a conjugate
pair, 
\be 
\label{eq:ccef} p(\beta | \alpha) &=& h(\beta) \exp\{\alpha^\top t(\beta) - a_g(\alpha) \},  \\
p(z_i, x_i | \beta) &=& h(z_i, x_i) \exp\{\beta^\top t(z_i, x_i) - a_l(\beta) \nonumber
\}.  \nonumber 
\ee 
The functions $t(\beta)=\langle \beta,-a_l(\beta)\rangle$ and $t(x_i,z_i)$ are the
sufficient statistics of the global hidden variables and of the local
contexts, respectively. The functions $a_g(\cdot)$ and $a_l(\cdot)$
are the corresponding log normalizers~\citep{bleistochastic}.  Annealed and tempered
variational inference apply more generally, but the CCEF allows us to analytically compute certain
expectations.

We derive AVI and VT simultaneously.  We
consider the annealed or tempered ELBO as a function
of the global variational parameters, 
\begin{align*}
 {\cal L}(\lambda;T) & \triangleq  {\cal L}(\lambda,\phi(\lambda;T);T),  \\
    \phi(\lambda;T) & \triangleq  \rm arg{}\max_{\phi} {\cal L}(\lambda,\phi;T). 
\end{align*}
We have eliminated the dependence on the local variational parameters by implicitly optimizing them in $\phi(\lambda;T)$. 
This new objective has the same optima for $\lambda$ as the original tempered or tempered ELBO. 
Following~\cite{bleistochastic}, the T-ELBO is
\begin{multline*}
{\cal L}_T(\lambda;T) =  {\mathbb E}_q[1/T_y] \textstyle{\sum_i}({\mathbb E}_q[t(x_i,z_i)] + \alpha) \nabla_\lambda a_g(\lambda) \\
			  -  \lambda^\top \nabla_{\lambda} a_g(\lambda)  + a_g(\lambda) - {\mathbb E}_q[\log C(T_y)] - {\mathbb E}_q[\log q(y)]. 
\end{multline*}
The annealed ELBO is obtained when replacing $\E_{q(y)}[1/T_{y}]$ by $1/T$
and dropping all other $y-$dependent terms.

\begin{algorithm}[t]
\caption{Annealed or tempered SVI}
\label{alg:svi_alg}
\begin{algorithmic}[1]
  \State Initialize $\lambda^{(0)}$ randomly. Initialize ${ T}>1$.
  \State Set the step-size schedule $\rho_t$.
  \Repeat
  \State Sample a data point $x_i$ uniformly. Compute its local variational parameters,
  \begin{center} $\phi_t = \frac{\mathbb{E}_{\lambda_{t}} [\eta_l (x_i^{(N)}, z_i^{(N)})]}{{ T}}.$\end{center}
  \State Compute the intermediate global parameters as if $x_i$ was replicated $N$ times, 
  \begin{center}$\hat{\lambda}_t = \alpha +\frac{1}{{T}} \mathbb{E}_{\phi_t} [ t(x_i^{(N)}, z_i^{(N)})]$. \end{center}
  \State Update the current estimate of the global variational parameters,
  \begin{center}
    $\lambda_{t+1} = (1 - \rho_t )\lambda_t + \rho_t \, \hat{\lambda}_t.$
  \end{center}
  \State {\bf Annealing}: reduce ${T}$ according to schedule. 
  \State {\bf Variational Tempering}: update ${T}$ with Eq.~\ref{eq:tupdate}/\ref{eq:tupdate2}.
  \Until{Forever}
\end{algorithmic}
\end{algorithm}

\subsection{Global and Local Random Variable Updates} 
To simplify the notation, let $1/T$ either be the deterministic inverse temperature for AVI, or ${\mathbb E}[1/T_y]$ for VT,
or a data~point-specific expectation ${\mathbb E}[1/T_i]$ as for LVT.
Following ~\cite{bleistochastic}, the 
natural gradient of the annealed ELBO with respect to the global variational parameters $\lambda$ is
 \begin{align*}
 \nabla_\lambda {\cal L} \; =\; \alpha + \sum_{i=1}^N \frac{1}{T}\mathbb{E}_q[t(x_i,z_i)] - \lambda. 
 \end{align*}
The variables $t(x_i,z_i)$ are the sufficient statistics from Eq.~\ref{eq:ccef}. Setting the gradient to zero
gives the corresponding coordinate update for the globals. Because of the structure of the gradient as
a sum of many terms, this can be converted into a stochastic gradient by subsampling from the data set,
$
\hat{\nabla}_{\lambda} {\cal L} = \alpha +\frac{1}{T} \mathbb{E}_q [t (x_i^{(N)}, z_i^{(N)})] - \lambda
$, where $t(x_i^{(N)}, z_i^{(N)} )$ are the sufficient statistics when data point $x_i$ is replicated $N$ times.
The gradient ascent scheme can also be expressed as the following two-step process,  
\be
\label{eq:global_update_}
\hat{\lambda}_t & = &  \alpha + \frac{1}{{T}} \, \E_q[{t(x_i^{(N)},z_i^{(N)})}], \n
\lambda_{t+1}  & = &(1 - \rho_t )\lambda_t + \rho_t \, \hat{\lambda}_t.  
\ee
We first build an estimate $\hat{\lambda}_t$ based on the sampled data point, and then merge this estimate
into the previous value $\lambda_t$ where $\rho_t$ is a decreasing learning rate. In contrast to SVI, we divide the expected sufficient statistics by temperature.
This is similar to seeing less data, but also reduces the 
variance of the stochastic gradient.

After each stochastic gradient step, we optimize the annealed or tempered ELBO over the locals.
The updates for the local variational parameters are
 \begin{align*}
 \phi_{nj} =\frac{1}{{T}} \mathbb{E}_q[{\eta_l (x_n,z_{(n,-j)},\beta)}] .
 \end{align*}
Above, $\eta_l$ is the natural parameter of the original (non-annealed) exponential family distributions of the local variational parameters~\citep{bleistochastic}.
As for the globals, the right hand side of the update gets divided by temperature. 
We found that tempering the local random variable updates is the crucial part in models that involve discrete variables. 
This initially softens the multinomial assignments and leads to a more uniform and better growth of the global variables.

\subsection{Updates of Variational Tempering}
\label{sec:updates_VT}
We now present the updates specific to VT.
In contrast to annealed variational inference, variational tempering
optimizes the \emph{tempered} ELBO, Eq.~\ref{eq:telbo}. 
As discussed before, the global and local updates of AVI are obtained from the global and local updates
of VT upon substituting $\E_{q(y)}[1/T_{y}] \rightarrow 1/T $.
Details on the derivation for these updates are given in the Supplement on the example of LDA.

The temperature update follows from the tempered ELBO.
To derive it, consider the log complete conditional for $y$ that is (up to a constant)
\begin{align*}
\log p(y_{m}| \cdot) & =  y_{m} \left(\frac{1}{T_m}\sum_i \log p(x_i,z_i | \beta) + \log \frac{\pi_m}{C(T_m)} \right).
\end{align*}
The variational update for
a multinomial variable is
\be
\label{eq:tupdate}
 r^*_{m}   \propto  \exp \left\{ \textstyle \frac{1}{T_m}\sum_i\E_q[\log p(x_i,z_i|\beta)]   + \log \frac{\pi_m}{C(T_m)}  \right\}. 
\ee

Let us interpret the resulting variational distribution over temperatures.
First, notice that the expected local likelihoods $\E[p(\bx,\bz|\beta)] $ enter the multinomial weights, 
multiplied with the vector of inverse temperatures. This way, small likelihoods (aka poor fits) favor distributions
that place probability mass on large temperatures, i.e. lead to a tempered posterior with large variances.  
The second term is the log tempered partition function, which is monotonically growing 
as a function of $T$. As it enters the weights with a negative sign, this term favors low temperatures. 

This analysis shows that the distribution over temperatures is essentially controlled by the likelihood: large
likelihoods lead to distributions over temperature that place its mass around low temperatures,
and vice versa. 
As likelihoods
increase, the temperature distribution shifts its mass to lower values of $T$. 
This way, the model controls its own annealing schedule.
Algorithm~\ref{alg:svi_alg} summarizes variational tempering.

\parhead{Estimation of the tempered partition function.}
Let us sketch how we can approximate the normalization constants $C(T_m)$ for a discrete set of $T_m$. 
At first sight, this task might seem difficult due to the high dimensionality of the joint. But note that
in contrast to the posterior, the joint distribution is highly symmetric, and therefore calculating its normalization is tractable.

In the supplement we prove the following identity for the considered class of CCEF models,
\be
C(T_m) =  \int d\beta\, p(\beta) \prod_{i=1}^N  \left( \int\,dz_i\,dx_i\, e^{ (\beta^\top t(x_i,z_i) - a_l(\beta))/T_m }\right) \n
	 \stackrel{(A.2)}{=}  \int d\beta\, p(\beta)  \exp\{ - N  a_l(\beta)/T_m +  N a_l(\beta/T_m)  \}. \nonumber
\ee
The dimension of the remaining integral is independent of the size of the data set; it is therefore of much lower dimension
than the original integral. We can therefore approximate it by Monte-Carlo integration. We found that $100$ samples are typically enough,
each integral typically takes a few seconds in our application. We can alternatively also replace the integral by a MAP approximation (see Supplement A.2).
Note that the normalization constants can be precomputed.

\subsection{Updates of Local Variational Tempering}
For multinomial local temperature variables, the updates are given in analogy to Eq.~\ref{eq:tupdate}:
\be
r^*_{m,i}  \; \propto \;  \exp \left\{ \textstyle \frac{1}{T_m} \E_q[\log p(x_i,z_i|\beta/T_m) + \log \pi_m] \right\}. \label{eq:tupdate2}
\ee
Thus, the likelihood's parameter gets divided by $T$.

In local variational tempering, the global variable is not conjugate due to temperature-specific sufficient statistics.
For mean-field variational inference, the variational distribution over the globals would need sufficient statistics of
the form $t(\beta) = \left<\beta,-a_l(\beta/T_1),-a_l(\beta/T_2), \cdots, -a_l(\beta/T_m)\right>$ which is not conjugate to the model prior,
thus we approximate to create a closed-form update.

The first sufficient statistics $\beta$ is shared between our chosen  variational approximation and the
optimal variational update for the locally tempered model. The second parameter in the variational approximation
scales with the number of data points. In the optimal variational update, these data points get split across 
temperatures. As an approximation, we assign these all to temperature $1$. This results in the following variational update:
\be
\hat{\lambda} & = & \alpha +  \E[1/T_i] \E[t(x_i^{(N)},z_i^{(N)})].
\ee
This looks the same as the first line in Eq.~\ref{eq:global_update_}
but assigns a different weight to each data point. 
Thus, in short, we match the first component of the sufficient statistics and pretend it came from a non-tempered model.

\begin{figure*}
\begin{center}
  \includegraphics[width=.32\linewidth]{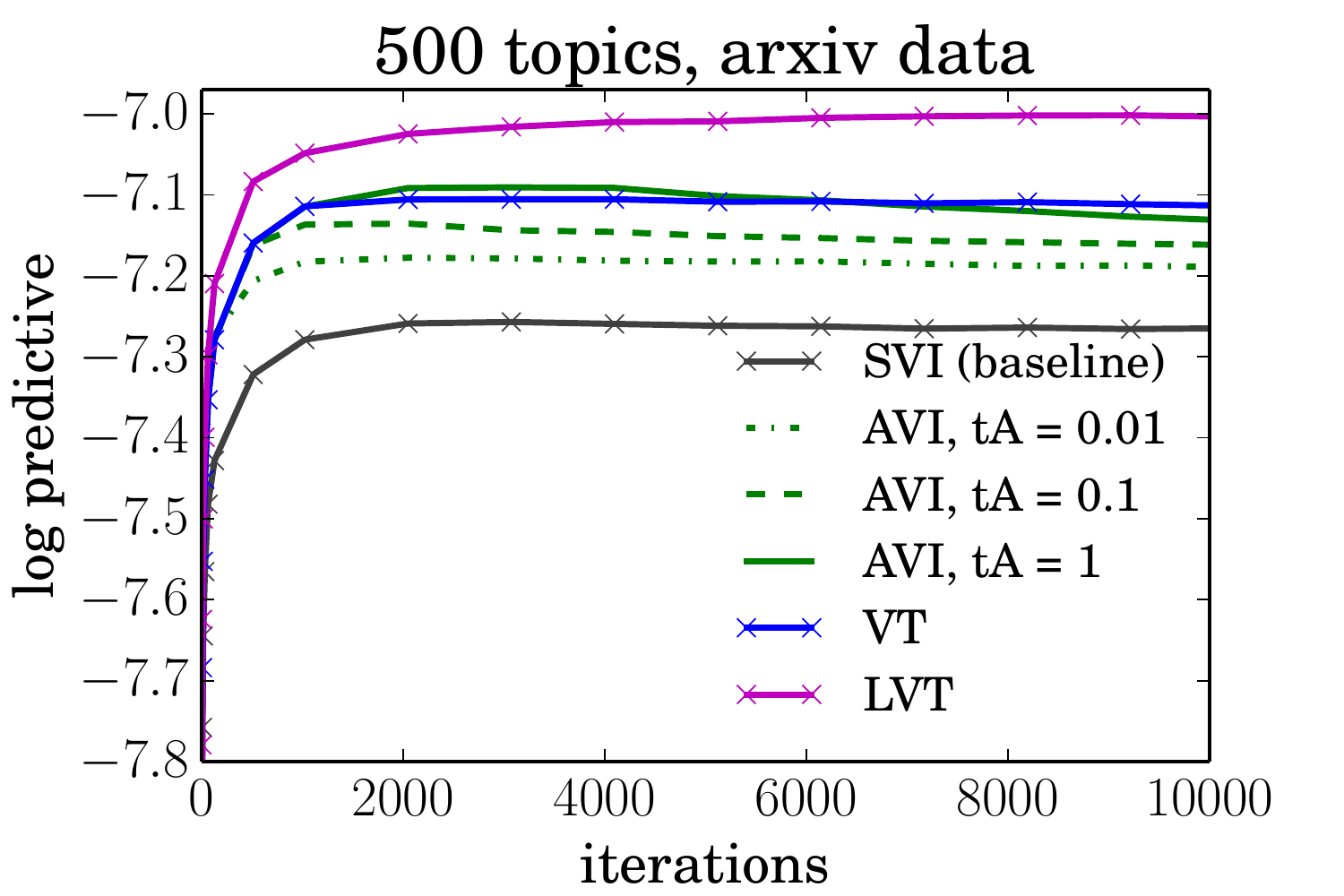}
  \includegraphics[width=0.32\linewidth]{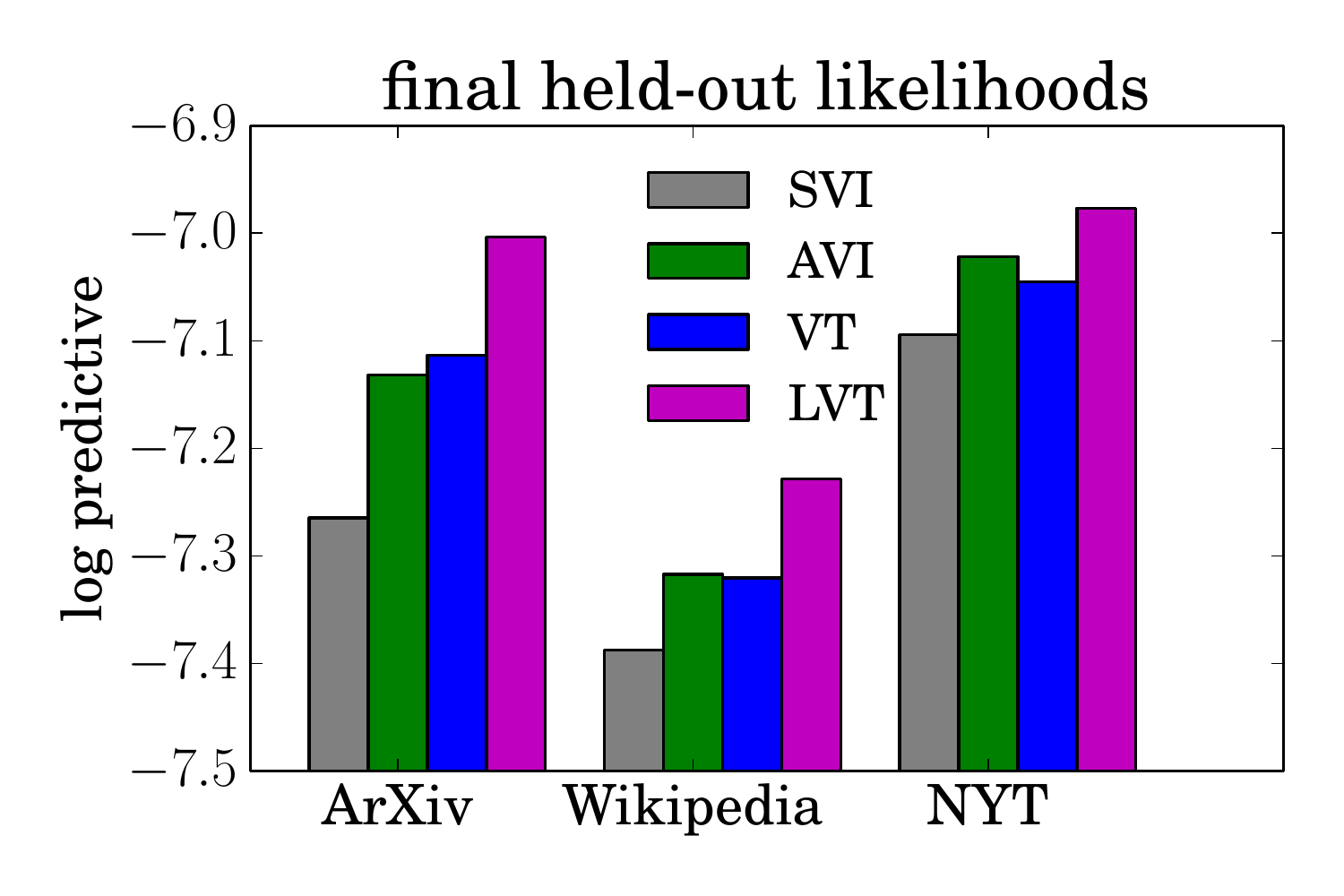}
  \includegraphics[width=.32\linewidth]{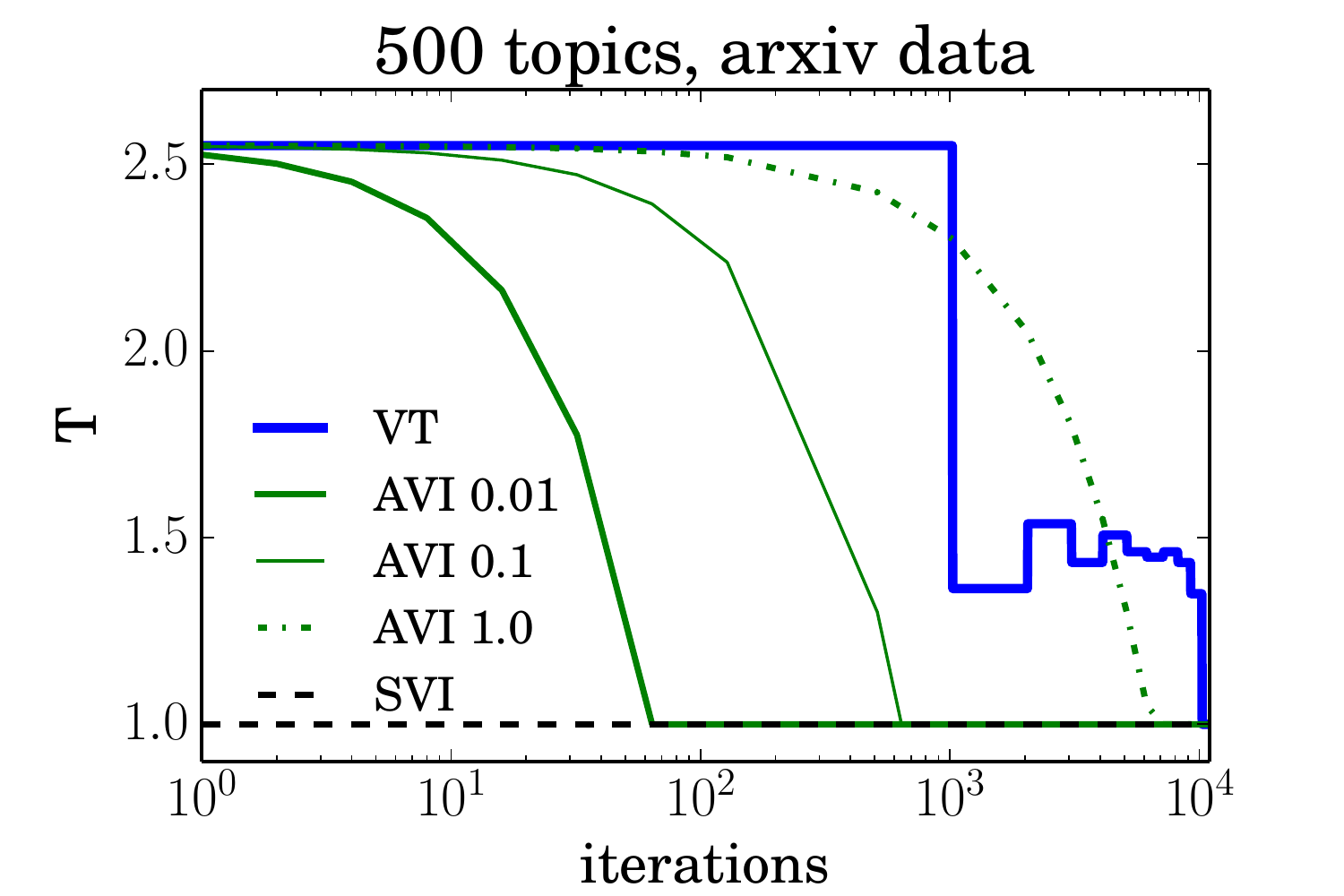}
  \caption{Log predictive likelihoods. We compare SVI~\citep{bleistochastic}, against VT (this paper),  LVT (this paper) and
     AVI (this paper) for different temperature schedules ($tA$ = length of the annealing schedule
    in effective traversals of the data). 
    Right: temperature schedules for the ArXiv data set. 
     }
\label{fig:lda}
\end{center}
\end{figure*}

\section{Empirical Evaluation}

For the empirical evaluation of our methods, we compare our
annealing and variational tempering approaches to standard SVI with latent
Dirichlet allocation on three massive text corpora. 
We also study batch variational inference, AVI, and VT on a factorial mixture model on simulated data
and on 2000 images to find latent components.
We consider the held-out predictive log likelihood of the
approaches~\citep{bleistochastic}. To show that VT and LVT
find better local optima in the original variational objective, 
we predict with the non-tempered model.\footnote{Even higher likelihoods were obtained 
using the learned temperature distributions for prediction, implying a beneficial role for the tempered model in dealing with outliers and model mismatch, similar to~\cite{mcinerney2015population}. We use the non-tempered model to isolate the optimization benefits of tempering.}
Deterministic annealing provides a significant improvement over standard SVI. 
VT performs similarly to the best annealing schedule and further inspection of the automatically
learnt temperatures indicate that it approximately recovers the best
temperature schedule. LVT (which employs local temperatures) outperforms
VT and AVI on all text data sets and was not tested on the factorial model.

\parhead{Latent Dirichlet allocation.} We apply all competing methods
to latent Dirichlet allocation (LDA) \citep{blei2003latent}. LDA is a
model of topic content in documents. It consists of a global set of
topics $\beta$, local topic distributions for documents $\theta_d$,
words $w_{dn}$, and assignments of words to topics $\bz$. Integrating
out the assignments ${\bf z}$ yields the multinomial formulation of
LDA $ p(w,\beta,\theta) = \textstyle p(\beta) p(\theta) \prod_{nd}
\left(\sum_k \theta_{dk}\beta_{kw_{dn}}\right)$.
Details on LDA can be found in the Supplement.

\parhead{Datasets.} We studied three datasets: 1.7 million articles
collected from the {\bf New York Times} with a vocabulary of 8,000
words; 640,000 {\bf arXiv} paper abstracts with a vocabulary of 14,000
words; 3.6 million {\bf Wikipedia} articles with a vocabulary of 7,702
words. We obtained vocabularies by removing the most and least
commonly occurring words.

\parhead{Hyperparameters and schedules.} We used $K=500$ topics and
set $\eta$ and $\alpha$ to $1/K$ (we also tested different
hyperparameters and found no sensitivity). Larger topic numbers make
the optimization problem harder, thus yielding larger improvements of
the annealing approaches over SVI. We furthermore set batch size
$B = 100$ and followed a Robbins-Monro learning rate with
$\rho_t = (\tau + t)^{-\kappa}$, where $\tau=1024$, $\kappa=0.7$ and
$t$ is the current iteration count (these were found to be optimal
in~\citep{adaptive}). For SVI we keep temperature at a constant 1.
For VT, we distributed $100$ temperatures $1 \leq T_m \leq 10$ on an exponential scale and initialized $q(y_n)$ uniformly over the $T_m$. 
We precomputed the tempered partition functions $C(T_m)$ as discribed in the Supplement.
For annealing, we used linear schedules
that started in the mean temperature under a uniform distribution over $T_m$,
and then used a linearly decreasing annealing schedule that ended in $T=1$ after $tA \in \{0.01,0.1,1\}$ effective passes.
We updated $T$ every $1000$ iterations. For LVT, we employed $100$ per-document inverse temperatures evenly spaced between $0$ and $1$.

\parhead{Results} We present our results for annealing and
variational tempering. We test by comparing the predictive log
likelihood of held out test documents. We use half of the words in
each document to calculate the approximate posterior distribution over
topics then calculate the average predictive probability of the
remaining words in the document (following the procedure outlined in
\citep{bleistochastic}).

Figure~\ref{fig:lda} shows predictive performance. 
We see that annealing significantly improves predictive likelihoods with respect to SVI across datasets. 
In the plot, we index temperature schedules by \emph{tA}, indicating the number of passes through the dataset. 
Our results indicate that slow annealing approaches work better (\emph{tA}=1 is the best performing annealing curve). 
VT automatically chooses the temperature schedule and is able to recover or improve upon the best annealing curve for arXiv and the New York Times. 
Variational tempering for Wikipedia is close to that of the best annealing rate, and better than several other manual choices of temperature schedule.
In all three cases, LVT gives significantly better likelihoods than AVI and VT.

\begin{figure*}
  \includegraphics[width=.32\linewidth]{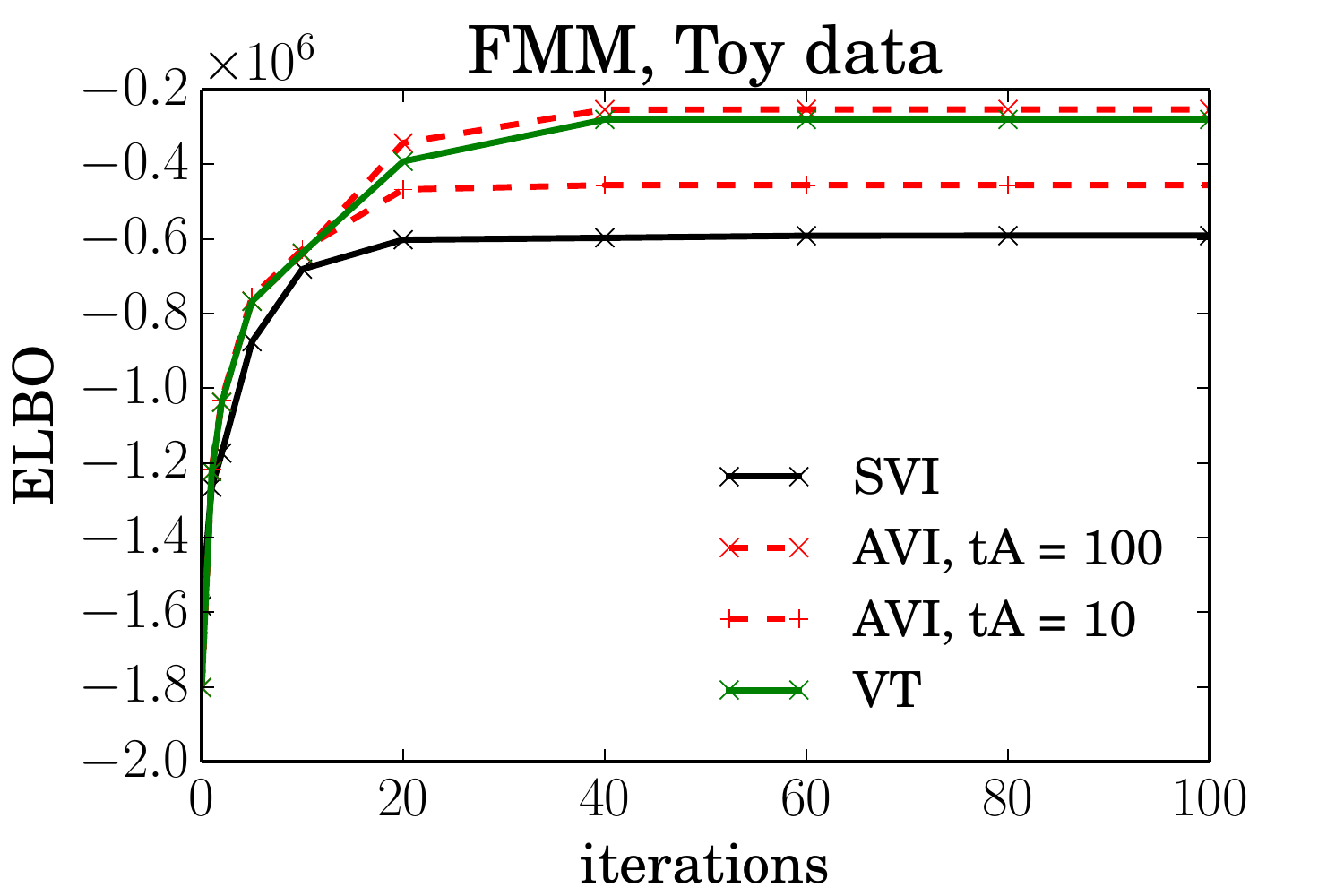}
  \includegraphics[width=.32\linewidth]{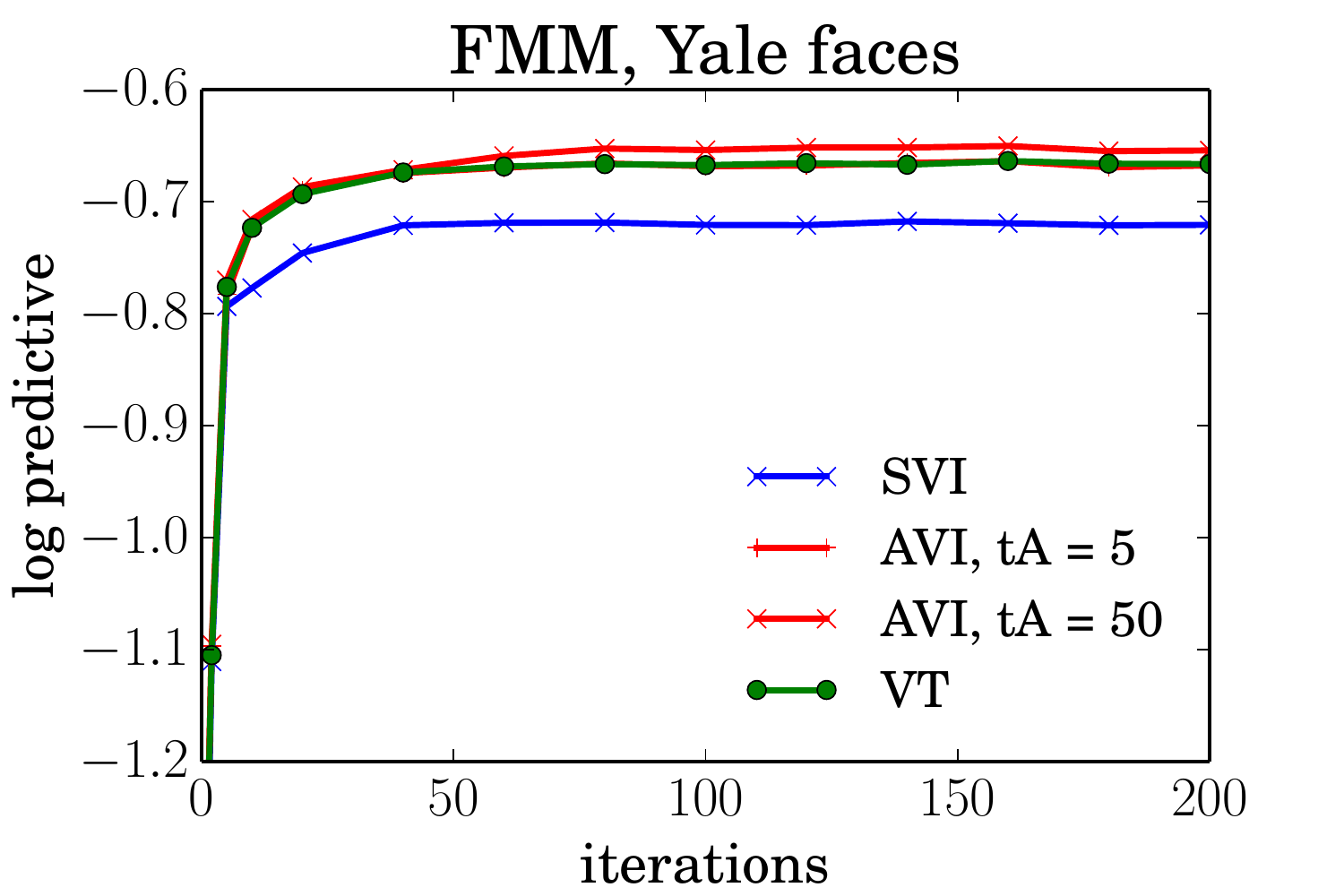}
  \includegraphics[width=.32\linewidth]{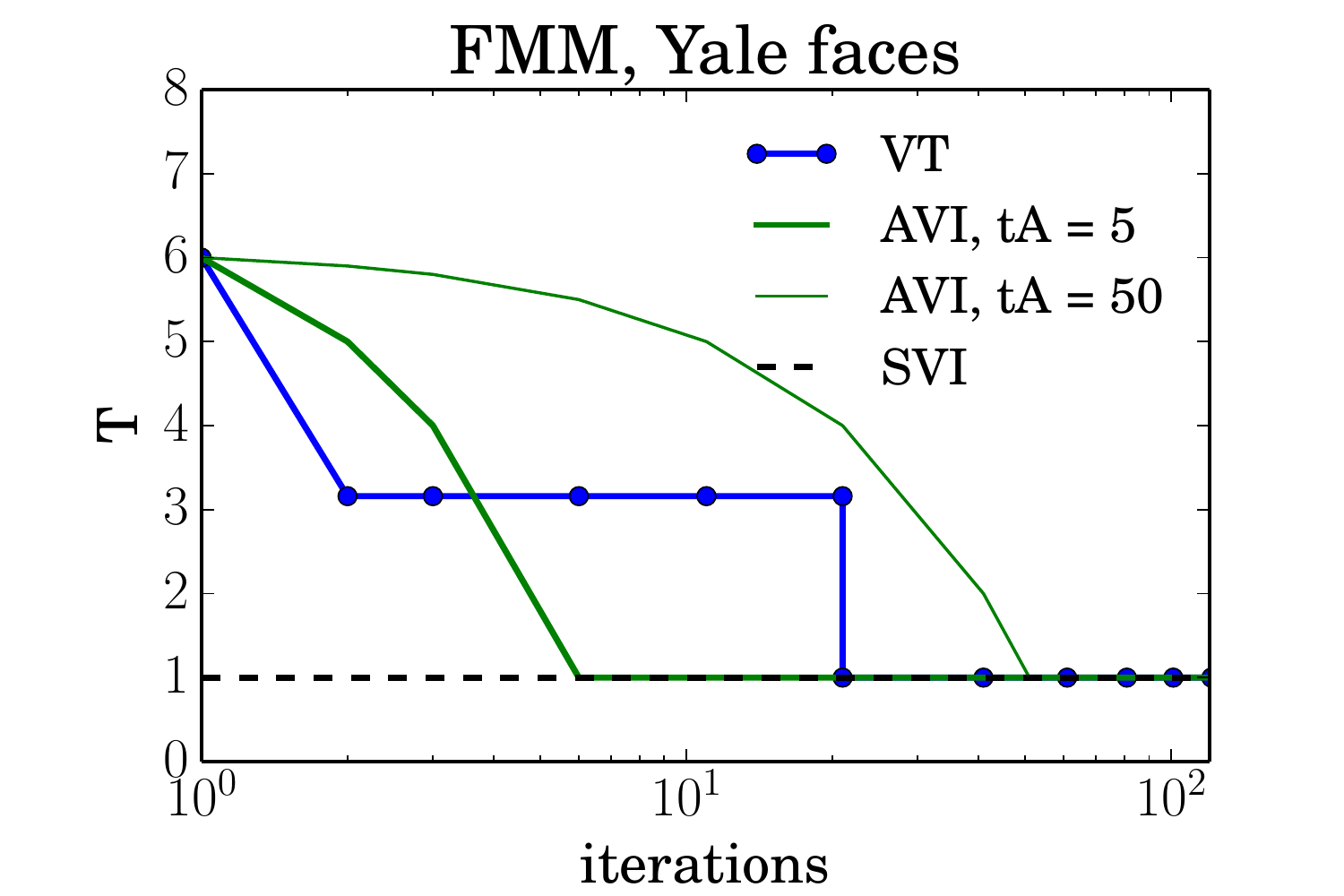}
  \caption{Factorial mixture model (FMM).
   { Left:} Evidence lower bound (ELBO) of toy data at $T=1$ for VI~\citep{jordan1999introduction}, annealed VI (AVI, this paper) and variational tempering (VT, this paper). 
   {Middle}: Log predictive likelihoods on Yale faces. {Right}: Expected temperatures on Yale faces, as a function of iterations for VI, AVI (linear schedules) and VT. 
   }
\label{fig:ibp_yale}
\end{figure*}

\parhead{Factorial mixture model.} 
We also carried out experiments on the Factorial Mixture Model (FMM)~\citep{ghahramani1995factorial,doshi2009variational}. 
The model assumes $N$ data points ${\bf X}_n\in {\mathbb R}^D$,
 $K$ latent  components $\mu_k\in {\mathbb R}^D$, and a $K \times N$ binary matrix of latent assignment variables ${\bf Z}_{nk}$.
The model has the following generative process~\citep{doshi2009variational}:
\be
{\bf X}_n  \; = \; \sum_k {\bf Z}_{nk} {\bf \mu}_k + {\bf \epsilon}_n, \quad {\bf Z}_{nk}\sim {\rm Bern}(\pi_k), \\
  \mu_k \sim {\cal N}(0,\sigma_\mu), \quad \epsilon_n \sim {\cal N}(0,\sigma_n). \nonumber
\ee
The variables ${\bf Z}_{nk}$ indicate the activation of factor $\mu_k$ in data point $n$.
Every ${\bf Z}_{nk}$ is independently $0$ or $1$, which makes the model different from 
the Gaussian mixture model with categorical cluster assignments.
The factorial mixture model is more powerful, but also harder to fit.

We are interested in learning the global variables $\mu_k$. 
We show in the Supplement that the log partition function for the factorial mixture model is
\be
\log C(T) & = & \frac{1}{2} ND \log T + N \sum_{k=1}^K \log (\pi_k^{1/T} + (1-\pi_k)^{1/T}). \nonumber
\ee
For details on the inference updates, see~\citep{doshi2009variational}.

\begin{figure}
\begin{center}
  \includegraphics[width=0.28\linewidth]{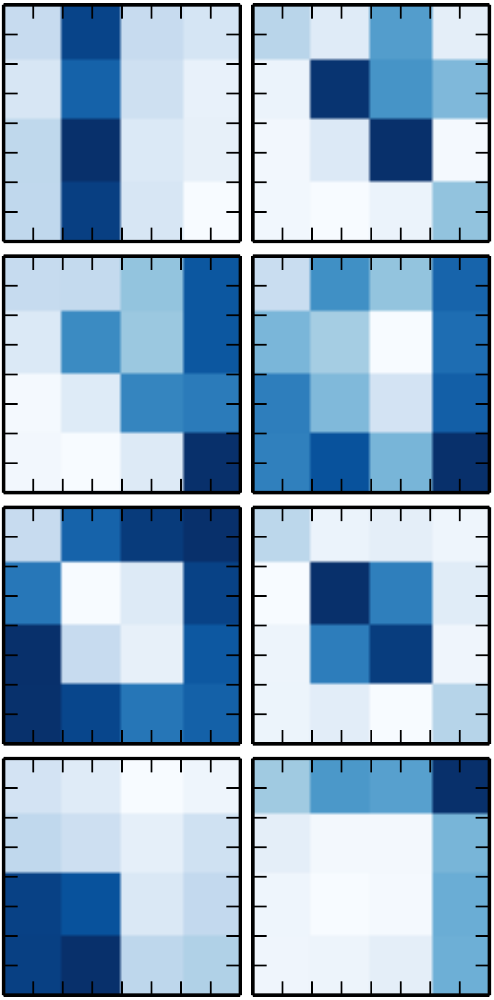}
\hspace{0.01\linewidth}
  \includegraphics[width=0.28\linewidth]{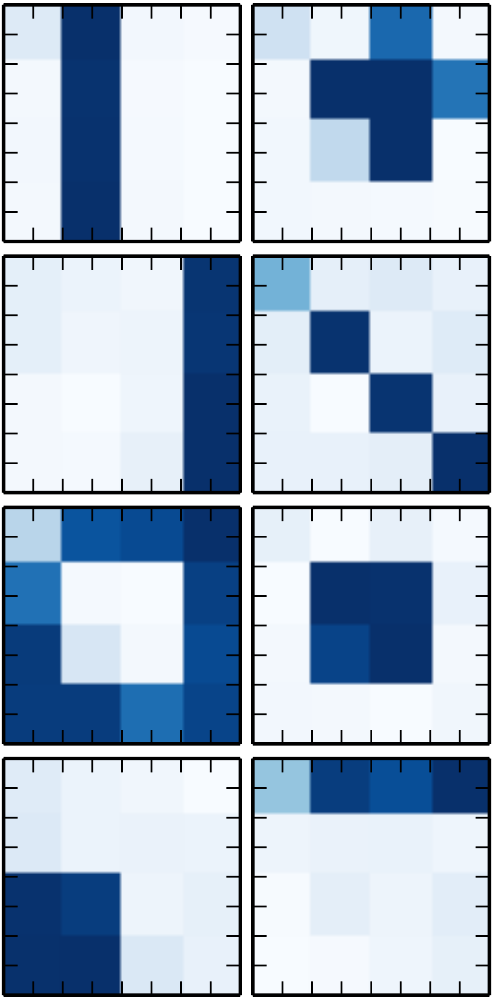}
\hspace{0.01\linewidth}
  \includegraphics[width=0.28\linewidth]{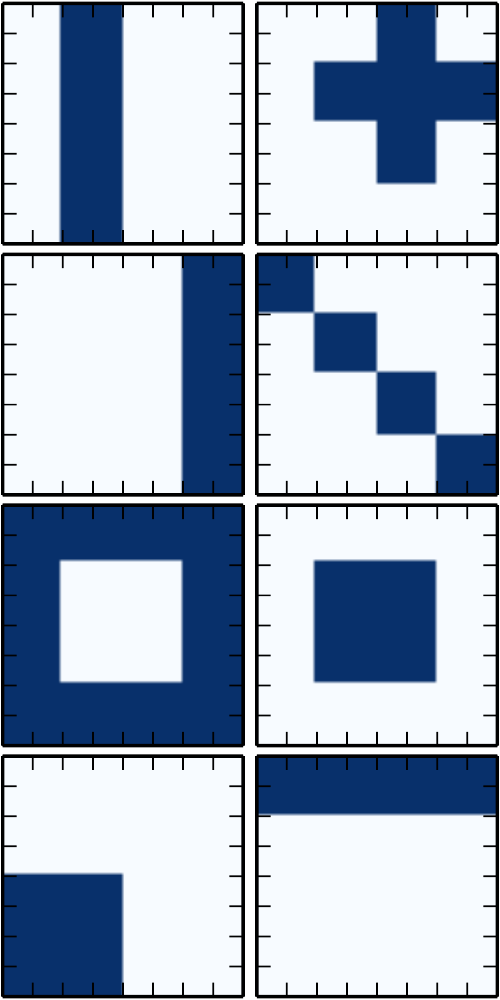} \n
(a) \quad  \quad  \quad \quad \quad\quad(b) \quad\quad \quad  \quad \quad \quad(c)
 \caption{FMM on toy data.
The shapes on the bottom show the latent global variables as found by (a) variational inference~\citep {jordan1999introduction} and (b) variational tempering (this paper).  Figure (c) shows the ground truth that was used to generate the data.}
\label{fig:ibp_toy}
\end{center}
\end{figure}

\parhead{Datasets and hyperparameters.} We carried out experiments two data sets. The first artificial data set that was generated by first creating 
$8$ components $\mu_k$ by hand. These are $4\times 4$ black and white images, i.e. binary matrices, each of which we weighted with a uniform draw from $[0.5,1]$.
(These are shown in Fig.~\ref{fig:ibp_toy}.) Given the $\mu_k$, we generated 10,000 data points from our model 
with $\sigma_n = 0.1$, and $\pi_k=0.3$. We set $\sigma_\mu=0.35$. Our linear annealing schedule started at $T=10$ and ended in $T=1$ at $10$ and $100$ iterations, respectively.
The second data set contained 2000 face images (Yale Face Database B, cropped version\footnote{http://vision.ucsd.edu/~leekc/ExtYaleDatabase/ExtYaleB.html}) from 
28 individuals with $168\times 192$ pixels in different poses and under different light conditions~\citep{lee2005acquiring}.
We normalized the pixel values by subtracting the mean and dividing by the standard deviation of all pixels. We chose $\sigma_n = \sigma_\mu = 0.5$ which we found to perform best.
Our annealing curves start at $T=6$ and end in $T=1$ after $5$ and $50$ iterations, respectively. Since both datasets were comparatively small,
we used batch updates.

\parhead{Results.}
Fig.~\ref{fig:ibp_toy} shows the results when comparing variational inference, annealed VI, and VT on the artificial data.
The left plot shows the ELBO at $T=1$. As becomes apparent, AVI and VT converge
to better local optima of the original non-tempered objective. The plots on the right are the latent features $\mu_k$
that are found by the algorithm. Variational tempering finds much cleaner features that agree better with the ground
truth than VI, which gets stuck in a poor local optimum. Fig.~\ref{fig:ibp_yale} shows held-out likelihoods for the Yale faces dataset.
Among the 2000 images, 500 were held out for testing. VT automatically finds an annealing schedule that comes close to the 
best linear schedule that we found. The plot on the right shows the different temperature schedules. 

\section{Conclusions}
We presented three temperature based algorithms for variational inference: annealed variational inference and global variational tempering, and local variational tempering. All three algorithms scale
to large data, result in higher predictive likelihoods, and
can be generalized to a broader class of models using black box variational methods.
VT requires model-specific precomputations
but results in near-optimal global temperature schedules. As such, AVI and LVT may be easier to use.
An open problem is to characterize and avoid new local optima
that may be created when treating temperature as a latent variable in variational inference.

\paragraph{Acknowledgements.}
This work is supported by NSF IIS-0745520, IIS-1247664, IIS-1009542, ONR N00014-11-1-0651, DARPA FA8750-14-2-0009, N66001-15-C-4032, Facebook, Adobe, Amazon, NVIDIA, and the Seibel and John Templeton
Foundations.

\bibliographystyle{apa} 
\bibliography{ref}

\newpage

\appendix

\section*{Supplementary Material}

\section{Tempered Partition Functions}

Variational Tempering requires that we precompute the tempered partition functions for
a finite set of pre-specified temperatures:
\be
C(T) & = & \int d\beta p(\beta) \prod_{i=1}^N \left(\int dx_idz_i p(x_i,z_i|\beta)^{1/T} \right).
\ee
We first show how to reduce this to an integral only over the globals. Because the size of the remaining integration
is independent of $N$,  it is tractable by Monte-Carlo integration with a few hundred to thousand samples.

\subsection{Generic model.}
\label{app:tempered_partition_function}
Here we consider a generic latent variable model of the SVI class, i.e. containing local and global hidden variables.
The following calculation reduces the original integral over global and local variables to an integral over the global
variables alone:
\be
\label{eq:C_NT}
C(T) & = & \int d\bx\,d\bz\,d\beta\, p(\beta)p(\bz,\bx|\beta)^{1/T} \\
		&  = &\int d\beta\, p(\beta) \prod_{i=1}^N \left(\int dx_i\, dz_i\, p(x_i,z_i| \beta)^{1/T} \right) \n
		  &  =  & \int d\beta\, p(\beta) \prod_{i=1}^N  \left( \int\,dz_i\,dx_i\,  \right. \n
		& & \left. \quad \quad \quad \times  \exp\{ \tinv \beta t(x_i,z_i) - \tinv a_l(\beta) \}\right). \nonumber
\ee
We now use the following identity:
\be
	&  &  \int\,dz_i\,dx_i\, \exp\{ \tinv \beta t(x_i,z_i) - \tinv a_l(\beta) \} \n
	& = &  \int\,dz_i\,dx_i\, e^{ \; \tinv \beta t(x_i,z_i) - \tinv a_l(\beta) + a_l(\tinv \beta) - a_l(\tinv \beta) } \n
	& = &  e^{ - \tinv a_l(\beta) + a_l(\tinv \beta) }  \underbrace{ \int\,dz_i\,dx_i\,e^{ \; \tinv \beta t(x_i,z_i) -a_l(\tinv \beta)} }_{=1} \n
		   & = &   \exp\{ -   \tinv a_l(\beta) +   a_l(\tinv \beta)  \}. \nonumber
\ee
Note that the integral is independent of $i$, as all data points contribute the same amount to the
tempered partition function. Combining the last 2 equations yields
\be
C(T) & = &  \int d\beta\, p(\beta)   \exp\{ -  N \tinv a_l(\beta) +  N a_l(\tinv \beta)  \}. \nonumber
\ee
The complexity of computing the remaining integral does not depend on the number of data points, and therefore
it is tractable with simple Monte-Carlo integration. We approximate the integral as
is
\be
\log C(T)  \approx  \log \frac{1}{N_s}\sum_{\beta \sim p(\beta)} \exp \left\{  - N \tinv a_l(\beta) +  N a_l(\tinv \beta)     \right\}. \nonumber
\ee
For the models under consideration, we found that typically less than 100 samples suffice. 
In more complicated setups, more advances methods to estimate the Monte Carlo integral can be used,
such as annealed importance sampling.
While we typically precompute the tempered partition function for about 100 values of T,
the corresponding computation could easily be incorporated into the variational tempering algorithm.

\paragraph{Analytic approximation.} Instead of precomputing the log partition function, one could also use an analytic approximation
for sparse priors with a low variance (this approximation was not used in the paper). 
In this case, we can MAP-approximate the $\beta-$integral, which results in
\be
\log C(T) & \approx & N \left(  T^{-1}  a_{l}(\beta^*) - a_l(T^{-1}\beta^*) \right).
\ee
For large data, this approximation gets better and yields an analytic result.

\section{Latent Dirichlet Allocation}
\label{sec:eg_lda}
\subsection{Tempered partition function}

We now demonstrate the calculation of the tempered partition function on the example of Latent Dirichlet Allocation (FFM).
We use the multinomial representation of LDA where the topic assignments are integrated out,
\be
p(w,\beta,\theta) =  p(\beta) p(\theta) \prod_{nd}  \left(\sum_k \theta_{dk}\beta_{kw_{dn}}\right). \label{eq:full_distrib}
\ee
The probability that word $w_{dn}$ is the multinomial parameter $\sum_k \theta_{dk}\beta_{kw_{dn}}$.
In this formulation, LDA relates to probabilistic matrix factorization models.

LDA uses Dirichlet priors $p(\theta)  =  \prod_d \text{Dir}(\theta_d | \alpha)$ and $p(\beta)  =  \prod_k \text{Dir}(\beta_k | \eta)$
for the global variational parameters $\beta$ and the per-document topic proportions $\theta$.

The inner "integral" over $w_{dn}$ is just the sum over the multinomial mean parameters,
\be
 \int d {w_{dn}} \; p(w_{dn} |\theta_d,\beta)^{1/T} & = & \sum_{v=1}^V   \left( \sum_k \theta_{dk}\beta_{kv} \right)^{1/T}.
\ee
The tempered partition function for LDA is therefore
\begin{align}
C(T) & =  \int d\beta p(\beta)  \prod_{d=1}^D \int d\theta_d p(\theta_d)  \left(\sum_v   \left( \sum_k \theta_{dk}\beta_{kv} \right)^{1/T}\right)^{N_d} \n
	& \approx  \int d\beta p(\beta)\left( \int d\theta \,p(\theta)  \left(\sum_v   \left( \sum_k \theta_{k}\beta_{kv} \right)^{1/T}\right)^{N} \right)^D, \nonumber
\end{align}
where as usual $N = W/D$ is the approximate number of words per document.
The corresponding Monte-Carlo approximation for the log partition function is
\be
\log C(T) &\approx& \log \frac{1}{N_\beta}\sum_{\beta \sim p(\beta)}\exp \{D \log \frac{1}{N_\theta} \\
&& \times\sum_{\theta \sim p(\theta)} \exp (N \log \sum_{v}(\sum_k \theta_k \beta_{kv})^{1/T}))\}.\nonumber
\ee
$N_\theta$ and $N_\beta$ are the number of samples from $p(\theta)$ and $p(\beta)$, respectively.
To bound the log partition function we can now apply Jensen's inequality
twice: Once for the concave logarithm, and once in the other direction for the convex
functions $x\rightarrow x^D$ and $x\rightarrow x^N$:
\be
\log C(T) \geq N\cdot D\int d\beta p(\beta )\int d\theta p(\theta ) \log \sum_v   \left( \sum_k \theta_{k}\beta_{kv} \right)^{1/T}, \n
\log C(T) \leq N\cdot D \log \int d\beta p(\beta )\int d\theta p(\theta ) \sum_v   \left( \sum_k \theta_{k}\beta_{kv} \right)^{1/T}. \nonumber
\ee
We see that the log partition function scales with the total number
of observed words $N\times D$,
This is conceptually important because otherwise $\log C(T)$ would have no effect on the updates in the limit of large data sets.

\subsection{Variational updates}
In its formulation with the local assignment variables $z_n$, the LDA model is
\be
p(w,z,\beta,\theta)  =  p(\beta) p(\theta)\prod_{n,d,k} \exp\{z_{dnk} (\log \theta_{dk} + \log \beta_{kw_{dn}}) \}. \nonumber
\ee
Let $N_{tot}$ be overall the number of words in the corpus, $N_{d}$ the number of words in document $d$, $D$ the number of documents, and $K$
the number of topics. We have that $\sum_{d=1}^D N_d = N_{tot}$. The tempered model becomes
\begin{multline*}
p(w,z,\beta,\theta,y)  =  p(\beta) p(\theta) \prod_{m}  \times \\
 \exp\{y_m \left[\sum_{n,d,k}z_{dnk} \frac{\log \theta_{dk} + \log \beta_{kw_{dn}}}{T_{m}} -  \log C(T_m) \right]\}. \nonumber
\end{multline*}
where $m$ indexes temperatures.
\paragraph{Variational updates.}
We obtain the following optimal variational distributions from the complete conditionals (all up to constants). 
We replaced sums over word indices $n$ by sums over the
vocabulary indices $v$, weighted with word counts $n_{dv}$:
\begin{align}
\log q^*(z_{dvk}) & =  z_{dvk} n_{dv} {\mathbb E}[1/T_{y}] ( {\mathbb E}[ \log \theta_{dk}] + {\mathbb E}[\log \beta_{kv}   ] ) , \\
\log  q^*(y_m) & =  y_{m} \left[ \frac{1}{T_m}\sum_{vk} n_{dv} \, {\mathbb E}[z_{dvk}] ({\mathbb E}[ \log \theta_{dk}]\right. \n
			& +\left.{\mathbb E} [\log \beta_{kv}]) - {\mathbb E}[\log C(T_y)] \right] , \n
\log  q^*(\theta_{dk}) & =  \log \theta_{dk} \left(\sum_{v}  n_{dv}\, {\mathbb E}[1/T_y]  {\mathbb E}[z_{dvk}] + \alpha \right) , \n
\log  q^*(\beta_{kv}) & =  \log \beta_{kv} \left(\sum_{d} n_{dv}\,  {\mathbb E}[ 1/T_y]  {\mathbb E}[z_{dvk}]   + \eta \right). \nonumber
\end{align}

\section{Tempered Partition function for the Factorial Mixture Model}
\label{sec:eg_mm}

We apply variational tempering to the factorial mixture model (FMM) as described in the main paper,
\be
p({\bf X},{\bf Z},\mu,\pi|\alpha,\mu_0) & = &  p({\bf X}|{\bf Z},\mu,\sigma_n) p({\bf Z}|\pi) p(\mu|\sigma_\mu).\nonumber
\ee
For convenience, we define the assigned cluster means for each data point:
\be
\mu_n({\bf Z}) &=& \sum_k {\bf Z}_{nk} \mu_k.
\ee
The data generating distribution for the FMM is now a product over $D-$dimensional Gaussians:
\be
p({\bf X}|{\bf Z},\mu,\sigma_n) & = &\prod_n {\cal N}({\bf X}_n; \mu_n({\bf Z}),\sigma_n{\bf I}_D) 
\ee
The local conditional distribution  also involves the prior $\prod_n p({\bf Z}_n|\pi)$ of hidden assignments,
\be
p({\bf X},{\bf Z}|\mu,\sigma_n) & = & \prod_n {\cal N}({\bf X}_n; \mu_n({\bf Z}),\sigma_n {\bf I}_D) p({\bf Z}_n|\pi).\nonumber
\ee
Here is the tempered local conditional:
\begin{multline*}
p({\bf X},{\bf Z} | \mu,\pi)^{1/T}   = \prod_n \frac{1}{\sqrt{ (2\pi  \sigma_n)^D }} \n
	\times \exp\left\{ -\frac{1}{2 \sigma_n T}\sum_n ({\bf X}_n - \mu_n)^\top ({\bf X}_n - \mu_n) + \frac{1}{T}\log p({\bf Z}_n| \pi)\right\}. 
\end{multline*}
When computing the tempered partition function, we need to integrate out all variables, starting with the locals.
We can easily integrate out ${\bf X}$; this removes the dependence on $\mu$ which only determines the \emph{means} of the Gaussians:
\be
C(T,{\bf Z},\pi) & = &\int d^D{\bf X}  \; p({\bf X},{\bf Z} | \mu,\pi)^{1/T}  \\
	     & = &  \left(\prod_{n=1}^N  \frac{\sqrt{  (2\pi \sigma_n T)^D  } }{\sqrt{  (2\pi\sigma_n )^D} } \right) \prod_{n} p({\bf Z}_n|\pi)^{1/T} \n
 	     & = & \sqrt{T}^{ND} \prod_{n} p({\bf Z}_n|\pi)^{1/T}. \nonumber
\ee
Hence, integrating the tempered Gaussians removes the $\mu-$dependence and gives an
analytic contribution $\sqrt{T}^{ND}$ to the tempered partition function. It remains to compute
\be
C(T) & = &\sqrt{T}^{ND} \int  d{\bf Z}  \prod_{n} p({\bf Z}_n|\pi)^{1/T}.
\ee
Since the Bernoulli variables are discrete, the last marginalization yields
\be
&&  \sum_{\{{\bf Z}_{nk}\}}\prod_{nk} \pi_k^{{\bf Z}_{nk}/T} (1-\pi_k)^{(1-{\bf Z}_{nk})/T}\n
  &=&  \prod_{nk}\sum_{\{{\bf Z}_{nk}=\pm 1\}} \pi_k^{{\bf Z}_{nk}/T} (1-\pi_k)^{(1-{\bf Z}_{nk})/T} \n
	& = & \prod_k \left(  \pi_k^{1/T} + (1-\pi_k)^{1/T}\right)^N.
\ee
Finally, the log tempered partition function is
\be
\log C(T) & = & \frac{1}{2} N D \log(T) + N \sum_k \log \left(  \pi_k^{1/T} + (1-\pi_k)^{1/T}\right) \n
		& = &  \frac{1}{2} ND \log(T) + N K   \log \left(  \pi^{1/T} + (1-\pi)^{1/T}\right) \nonumber.
\ee
In the last line we used that the hyperparameters $\pi_k \equiv \pi$ are isotropic in $K-$space.

\end{document}